\documentclass{article}
\usepackage[final]{colm2026_conference}

\usepackage{microtype}
\usepackage{hyperref}
\usepackage{url}
\usepackage{booktabs}
\usepackage{graphicx}
\usepackage{amsmath}
\usepackage{amssymb}
\usepackage{array}
\usepackage{float}
\usepackage{enumitem}
\usepackage{lineno}

\definecolor{darkblue}{rgb}{0, 0, 0.5}
\hypersetup{colorlinks=true, citecolor=darkblue, linkcolor=darkblue, urlcolor=darkblue}

\title{Reason-Mediated Behavioral Models for Auditing LLM Social Simulators}

\author{\normalfont
\begin{tabular}[t]{p{0.38\linewidth}p{0.38\linewidth}}
\textbf{Atharva Pandey} & \textbf{Gautam Jajoo}\\
Kairosity & Kairosity\\
\url{https://kairosity.ai} & \url{https://kairosity.ai}
\end{tabular}}

\newcommand{\D}{D}
\newcommand{\K}{K}
\newcommand{\X}{X}
\newcommand{\Z}{Z}
\newcommand{\Y}{Y}
\newcommand{\M}{M}
\newcommand{\Zhat}{\widehat{Z}}
\newcommand{\Yhat}{\widehat{Y}}
\newcommand{\Zcore}{\Z_{\mathrm{core}}}
\newcommand{\best}[1]{\textbf{#1}}
\newcommand{\second}[1]{\underline{#1}}

\begin{document}

\ifcolmsubmission
\linenumbers
\fi

\maketitle

\begin{abstract}
Large language models are increasingly used as social simulators, including as synthetic survey respondents. Most evaluations ask whether simulated outcomes resemble human outcomes. We argue that this is necessary but too weak: a simulator can match the final answer while using the wrong rationale-derived reason pattern. We study this problem through a 94-person sunscreen concept test in which each respondent evaluated three product concepts and wrote open-ended rationales. We map those rationales into signed reason states $\Z$, where positive signs support adoption and negative signs block it. This gives a practical audit: holding respondent descriptors $\D$, category context $\K$, and concept treatment $\X$ fixed, do human rationale-derived reasons help predict behavior $\Y$, and can an LLM simulate the same reason state without seeing the human rationale or outcome? Human rationale-derived reasons substantially improve held-out prediction of purchase intent. LLM-simulated reasons are more brittle: they often sound plausible, but frequently echo the concept board rather than recover the respondent's acceptance or rejection path. The paper contributes an evaluation framework for social simulators. Reason states do not identify natural causal effects by themselves, but they provide an interpretable test of whether a simulator's stated reasons align with human evidence.
\end{abstract}

\section{Introduction}

LLM-based social simulation is moving from demonstration to evaluation. Recent systems simulate individual agents, synthetic samples, and multi-agent societies~\citep{argyle2023out,park2023generative,anthis2025position}. The central question is no longer only whether these simulations are believable. It is whether they are faithful enough to support scientific or practical inference.

A common validation strategy compares simulated outcomes with human outcomes. In a concept test, for example, one might ask whether a model predicts the same winning product or a similar purchase-intent distribution. This is necessary, but not sufficient. Outcome agreement alone cannot tell us whether the model recovered the behavioral path that produced the outcome. A sunscreen concept can fail because it is too expensive, because its proof is not trusted, because the texture feels wrong for daily use, or because the respondent does not identify with the positioning. These mechanisms imply different interventions, even when they produce the same rating.

This paper studies a stricter object: the reason path between a stimulus and a response. We represent this path with a signed reason state $\Z$. A positive sign means that a reason supports adoption; a negative sign means that it blocks adoption; zero means that the reason is inactive. The goal is not to infer a natural causal effect from open-ended text alone. Rather, we ask whether a fixed, human-readable reason codebook can support a reason-mediated behavioral model for auditing social simulators.

The resulting test is simple. Respondent descriptors $\D$ include demographics, language, and personality items. Category context $\K$ contains sunscreen-specific prior state: use frequency, brands owned, white-cast concern, and trust in SPF claims. Concept treatment $\X$ contains the price, claims, authority cue, origin, and proof points in the concept board. Human rationales are mapped into $\Z$. A readout $f_\theta(\D,\K,\X,\Z)$ is trained to predict behavior $\Y$. We then replace human $\Z$ with LLM-simulated $\Zhat$ and ask whether that reason state helps the same readout. This separates two questions that are often conflated: can the model predict an answer, and can it recover the reason state that makes the answer intelligible?

We evaluate this protocol on a small sunscreen concept test with 94 human respondents and three concepts. Each respondent rated purchase intent, believability, and differentiation, wrote rationales, ranked concepts, and made a final pick. The study is intentionally narrow. Its purpose is to test the evaluation idea, not to make population claims about sunscreen buyers.

The paper makes three contributions.
\begin{enumerate}[leftmargin=1.4em, itemsep=2pt, topsep=2pt]
    \item We define a reason-mediated behavioral model for concept-test simulation: $\D,\K,\X$ shape $\Z$, and $\Z$ helps predict $\Y$.
    \item We show that human rationale-derived reason states improve held-out prediction of purchase intent.
    \item We show a negative result for current LLM reason simulation: LLM-simulated reasons can be fluent and plausible while failing to match the human reason path.
\end{enumerate}

\section{Related Work}

\paragraph{\textbf{LLM social simulation.}}
Prior work has shown that LLM agents can generate believable individual behavior, memories, plans, and social interactions~\citep{park2023generative}; synthetic samples can approximate some survey patterns~\citep{argyle2023out}; and agent societies can be used to explore group dynamics. At the same time, recent critiques emphasize that plausible simulation is not the same as valid social evidence. LLMs may flatten identity groups when used as participant replacements~\citep{wang2025misportray}, and apparent emergent dynamics may be observationally compatible with data leakage or prompt artifacts~\citep{barrie2025leakage}. These critiques motivate evaluations that inspect the mechanism of a simulated response, not only the surface form or final answer.

\paragraph{\textbf{Causal structure and refutation.}}
Causal inference papers often separate four tasks: modeling assumptions, identifying the estimand, estimating the effect, and refuting or stress-testing the result~\citep{sharma2020dowhy,pearl2009causality}. They also show why naive outcome comparisons can be misleading when the process that generates exposure is entangled with the process that generates behavior~\citep{sharma2015recommendation}. Work on LLMs and causality asks whether language models can produce valid causal arguments~\citep{kiciman2024causal}. Observed rationales are not randomized mediators, so they do not by themselves identify natural direct or indirect effects in the sense of mediation analysis~\citep{imai2010mediation}. A weaker but useful target is refutation: define an interpretable interface, hold the readout fixed, and check whether the interface behaves coherently under prediction, signed perturbation, ablation, and simulator substitution.

\paragraph{\textbf{Behavioral mediators.}}
The reason-state codebook is motivated by behavioral theories in which beliefs, attitudes, norms, values, and perceived constraints shape intentions and actions~\citep{fishbein1975belief,ajzen1991planned,schwartz1992universals}. For concept tests, the useful coding level is lower than broad constructs such as attitude or value. A node such as price/value, safety trust, or sensory fit is specific enough to be coded from text and manipulated in a model, but general enough to transfer across related concept tests.

\begin{figure}[t]
    \centering
    \includegraphics[width=0.88\linewidth]{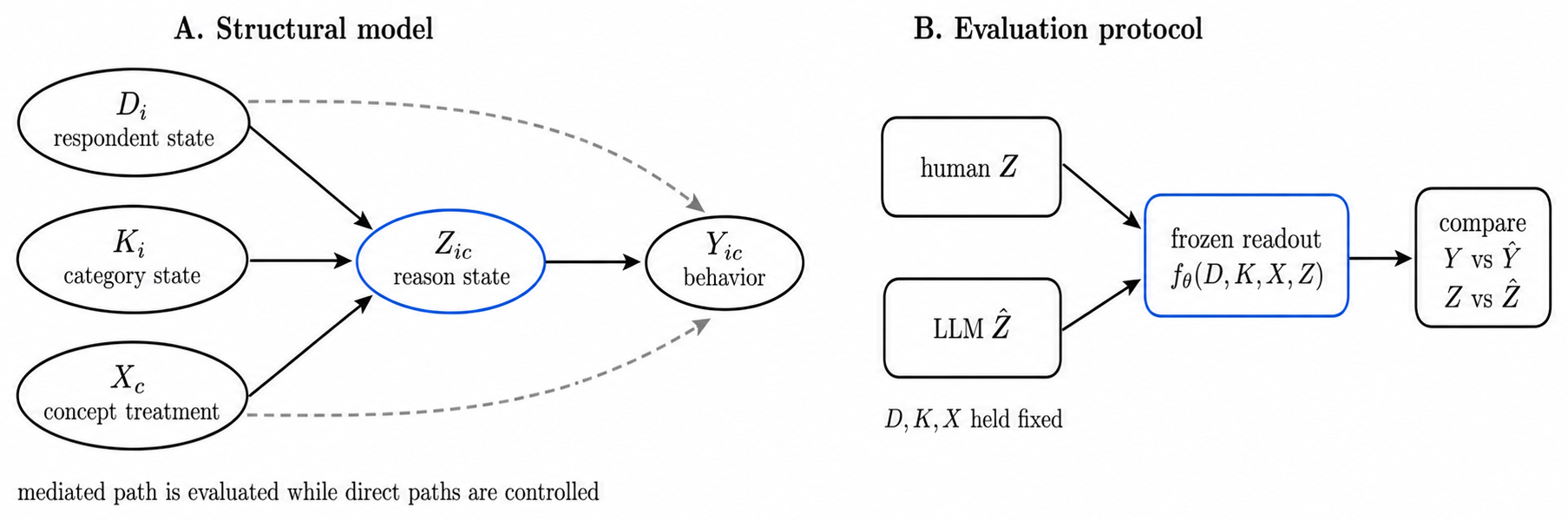}
    \caption{The reason-mediated behavioral model. We hold $\D,\K,\X$ fixed and test whether human or LLM-simulated reasons $\Z$ support the same behavioral readout. Bold arrows show the evaluated mediated path; dotted gray arrows show controlled direct paths.}
    \label{fig:graph}
\end{figure}

\section{Problem Setup}

\paragraph{\textbf{Variables.}}
The unit of analysis is a respondent-concept cell: respondent $i$ evaluating concept $c$. We use five objects:
\[
    \D_i,\quad \K_i,\quad \X_c,\quad \M_{ic},\quad \Z_{ic},\quad \Y_{ic}.
\]
$\D_i$ is the respondent state: demographics, language, and personality measures. $\K_i$ is the category state: sunscreen-use frequency, brands owned, white-cast concern, and trust in SPF claims. $\X_c$ is the concept treatment: price, claims, authority cue, positioning, and proof points. $\M_{ic}$ is the measured intermediate-rating vector, here believability and differentiation. $\Z_{ic}$ is the reason state expressed or predicted for respondent $i$ on concept $c$. $\Y_{ic}$ is behavior, mainly purchase intent on a 1--5 scale.

\paragraph{\textbf{Reason states.}}
$\Z$ is a signed vector of intermediate behavioral reasons:
\[
    \Z_{icj}\in\{-1,0,+1\}.
\]
For reason family $j$, $+1$ means the reason supports purchase, $-1$ means it blocks purchase, and $0$ means it is inactive. The sign is coded from the rationale, not from the purchase-intent score. Because the rationale is collected after the rating, $\Z$ is a post-rating rationale-derived representation rather than an observed pre-decision mediator. For example, price/value is positive when the respondent says the product is worth the price and negative when the respondent says it is too expensive for daily use.

\paragraph{\textbf{Core reason states.}}
The main paper uses a core reason-state vector, denoted $\Zcore$. This is the subset of reason families that are interpretable and not direct restatements of the outcome. It contains price/value, proof trust, natural orientation, premium or K-beauty aspiration, local fit, sensory/routine fit, and safety trust. Outcome-near labels, such as explicit purchase intention, are kept only for sensitivity checks because they can leak $\Y$.

\begin{table}[H]
    \centering
    \small
    \begin{tabular}{p{0.28\linewidth}p{0.25\linewidth}p{0.37\linewidth}}
        \toprule
        Reason node & Family & Sign meaning \\
        \midrule
        price/value & Value and constraint & $+1$: worth the price; $-1$: too expensive or poor value \\
        proof trust & Evidence and authority & $+1$: claims are trusted; $-1$: claims are doubted \\
        natural orientation & Natural/cultural cue & $+1$: natural or traditional cue helps; $-1$: cue reduces trust \\
        premium/K-beauty & Novelty and status & $+1$: premium or K-beauty appeal; $-1$: excessive or irrelevant \\
        local fit & Local adaptation & $+1$: fits Indian skin/climate; $-1$: local fit is missing \\
        sensory/routine fit & Daily-use fit & $+1$: texture/no-white-cast helps; $-1$: routine fit blocks use \\
        safety trust & Safety and risk & $+1$: safety is reassured; $-1$: safety/regulatory anxiety blocks use \\
        \bottomrule
    \end{tabular}
    \caption{Core $\Z$ codebook used in the main readout. The same node can be positive, negative, or inactive for a respondent-concept cell.}
    \label{tab:main-codebook}
\end{table}

\paragraph{\textbf{Reason-mediated behavioral model.}}
We use \emph{reason-mediated behavioral model} in a restricted, operational sense. The model represents behavior through an explicit reason state:
\begin{equation}
    p(\Y,\Z \mid \D,\K,\X)
    =
    p(\Y \mid \D,\K,\X,\Z)\;p(\Z \mid \D,\K,\X).
    \label{eq:factorization}
\end{equation}
This factorization should be read as an audit model rather than an identification claim for natural direct or indirect effects. Once a codebook and readout are fixed, we can ablate reason families, flip signs, and replace human reasons with LLM-simulated reasons. The question becomes whether the simulator places the right reason state in the middle, alongside getting the final answer right.

\paragraph{\textbf{Why direct outcome simulation is insufficient.}}
A direct synthetic respondent gives
\[
    \D,\K,\X \longrightarrow \Yhat.
\]
Our protocol asks for the stronger decomposition
\[
    \D,\K,\X \longrightarrow \Zhat \longrightarrow \Yhat.
\]
The second test can fail even when the final answer is directionally correct. That failure is useful: it catches plausible rationales that do not play the same behavioral role as human rationale-derived reasons.

\section{Methodology}

\subsection{Concept Test}

\paragraph{\textbf{Study.}}
The empirical setting is a sunscreen concept test with 94 usable human respondents and three product concepts, yielding 282 respondent-concept cells. The target sample was English-speaking Indian women aged 22--40, urban or semi-urban, with recent sunscreen purchase. Six responses were excluded for attention, rurality, or straight-line quality rules; the analytic sample had mean age 28.7 and city-tier counts of 51, 19, and 24 for Tier 1, Tier 2, and Tier 3. The three concepts varied on price, authority, cultural positioning, local fit, and sensory promise: SunVeda, an Ayurvedic/botanical concept at Rs. 499; Dr. Anita, a dermatologist/clinical Indian concept at Rs. 699; and Hae, a Korean prestige hybrid concept at Rs. 1,299. Appendix~\ref{app:study-details} gives the full distribution and concept text.

\paragraph{\textbf{Measured outcomes.}}
For each concept, respondents provided purchase intent, believability, differentiation, and an open-ended rationale. They then ranked the concepts and made a final pick. The main target is purchase intent $\Y\in\{1,\ldots,5\}$. We also report top-two-box intent, $T=\mathbb{1}[\Y\ge4]$, and final-pick checks.

\paragraph{\textbf{Questionnaire design.}}
The survey first collected screening and respondent descriptors, including age, city tier, education, income proxies, languages, and BFI-10 personality items. It then asked category questions about sunscreen-use frequency, brands owned, white-cast concern, and SPF-claim trust, before showing the three concept boards in randomized order. For each board, respondents answered purchase intent, believability, differentiation, and an open-ended ``why'' question. Thus $\D$ comes from respondent descriptors; $\K$ from sunscreen-category questions; $\X$ from the concept board; $\M$ from believability and differentiation; $\Z$ from the open-ended rationale; and $\Y$ from purchase intent, ranking, and final pick. Appendix~\ref{app:study-details} gives the full mapping and an example concept board.

\subsection{Reason-State Construction}

\paragraph{\textbf{Codebook.}}
The reason codebook maps open-ended rationales into signed reason states. It was built from the sunscreen concept-test task but uses families that are meant to be reusable across concept tests: value, proof trust, safety, local fit, novelty/status, routine fit, and category-specific authority cues. The leaves are domain-specific; the parent families are the transferable layer.

\paragraph{\textbf{Codebook construction.}}
The codebook was constructed before the predictive readout was evaluated. We first collected all open-ended rationales and listed recurring supports and objections without using purchase intent as a label. Surface variants were then merged into parent families, such as price/value, proof trust, local fit, safety trust, and sensory/routine fit. Outcome-near labels, including explicit intention or generic liking, were removed or kept only for sensitivity checks because they can leak $\Y$. Each retained family was assigned a signed coding rule: $+1$ for support, $-1$ for objection, and $0$ for inactive. The resulting main-paper codebook is Table~\ref{tab:main-codebook}; the appendix gives the machine-readable node names.

\paragraph{\textbf{Coding rule.}}
The extractor is allowed to use the respondent's rationale and the concept text for disambiguation, but not the rating, rank, final pick, or any gold reason label. It also cannot mark a reason only because the concept contains that attribute. For example, the Hae concept contains a K-beauty cue, but premium/K-beauty is coded positive only if the respondent uses that cue as a reason.

\paragraph{\textbf{Extractor.}}
The final human $\Z$ file was produced by a small-extractor, larger-verifier pipeline. A smaller LLM extractor (\texttt{gpt-5.4-mini}) coded the respondent-concept rationales into JSON over the fixed codebook, using the concept text only for disambiguation. Lexical and embedding extractors, including an Ollama \texttt{nomic-embed-text} run, were used as audit signals rather than as the final source of $\Z$. We then selected verifier rows where extraction was most likely to be brittle: zero-$\Z$ rows, warning cases, embedding--LLM disagreements, dense reason rows, and substantive hard cases such as Hae price rejection and Ayurveda skepticism. A larger verifier (\texttt{gpt-5.5}) recoded those rows; verifier outputs overrode the smaller-extractor outputs for the selected rows, while all other rows remained from \texttt{gpt-5.4-mini}. The adjudicated raw nodes were then compressed into the seven core $\Zcore$ families used in the main readout. No extractor saw purchase intent, believability, differentiation, ranking, final choice, or any gold reason label.

\subsection{Human Reason Model}

\paragraph{\textbf{Readout.}}
The human reason model is a supervised readout
\[
    f_\theta(\D,\K,\X,\Z)\rightarrow\Y.
\]
The primary readout is ridge regression with five-fold GroupKFold cross-validation by respondent, so all rows from a respondent remain in the same fold. Numeric variables are imputed and scaled; categorical variables are one-hot encoded; predictions are clipped to the 1--5 scale. This is the main model for purchase intent. We separately use balanced logistic regression to test whether the same features predict the respondent's final chosen concept. Random forests and gradient boosting are nonlinear checks: they test whether the main pattern depends on a linear readout.

\paragraph{\textbf{What this tests.}}
The human reason model asks whether human rationale-derived reason states have behavioral content after controlling for respondent descriptors, category context, and concept features. If $\Zcore$ improves held-out prediction, then it is more than decorative explanation.

\subsection{LLM Reason Simulation Protocols}

\paragraph{\textbf{Design.}}
We evaluate three pre-specified LLM simulation protocols. They answer different questions. \emph{Survey simulation} asks the LLM to behave like a respondent and produce answer text, which is then scored into a purchase-intent distribution. \emph{Reason-state simulation} asks the LLM to predict $\Zhat$ directly from $\D,\K,\X$ and the codebook. \emph{Behavior readout} asks whether a supplied reason state, human or simulated, predicts $\Y$ under the same human reason model. The comparison is therefore not one generic ``LLM simulation'' result; it separates final-answer simulation from reason-state simulation.

\paragraph{\textbf{Persona matching.}}
The persona-based survey simulations use a Nemotron-Personas-India pool~\citep{nvidia2025nemotronindia}. The matching code keeps female, urban personas aged 22--40 and assigns one persona to each human respondent using seeded stratified matching by city tier and age band. The primary bucket is $(\mathrm{city\ tier}, \mathrm{age\ band})$; fallbacks use same city tier or same age band. This is a prior over respondent background for persona-conditioned simulations. It should not be confused with observed $\D$: in the human study, $\D$ comes from survey fields; in persona-conditioned LLM simulations, part of the respondent description comes from the matched Nemotron persona.
The full simulation run includes demographic-only and persona-conditioned arms, so the persona prior can be ablated for native survey simulation. The main reason-state evaluation uses only arms that produce text from which a comparable $\Zhat$ can be extracted.

\paragraph{\textbf{Survey simulation: context-conditioned response.}}
The context-conditioned protocol uses Gemma 4 31B IT for high-volume synthetic respondent generation. It combines the matched persona with a stratum-level sunscreen context paragraph generated from non-outcome category variables only: sunscreen-use frequency, brands owned, white-cast concern, SPF-claim trust, languages, and personality summaries. Product outcomes and rationales are excluded. The model writes a short survey-style response. That response is scored into a purchase-intent probability mass function by semantic similarity rating (SSR)~\citep{maier2025ssr}, and the same generated text is mapped into $\Zhat$ for the $\Zhat\rightarrow\Y$ reason-path test. SSR mechanics are in Appendix~\ref{app:ssr}.

\paragraph{\textbf{Reason-state simulation: codebook-conditioned no-leak generator.}}
The codebook-conditioned protocol asks directly for $\Zhat$ using only observed $\D,\K,\X$ and the codebook. This constrained $\D,\K,\X\rightarrow\Zhat$ generation used the OpenAI verifier/generator pipeline rather than the Gemma survey-simulation model, because it is lower-volume and requires structured sparse JSON rather than respondent-style text. It does not use a Nemotron persona prior, and it does not see human open text, ratings, ranks, final pick, or gold reason labels. This is the strictest generation setting: the model must infer a sparse reason state from observed context and concept treatment alone. We therefore treat the results as protocol comparisons under a shared human reason model, not as a head-to-head model-family benchmark.

\subsection{Voice-Pool Survey Simulation}

\paragraph{\textbf{Motivation.}}
Single-prompt synthetic respondents tend to compress toward moderate answers. That is a problem for concept tests because many business decisions depend on tails: strong rejection due to price or safety doubt, and strong adoption due to proof, fit, or aspiration. The voice-pool simulation is a capacity test for this missing tail behavior. For the same respondent-concept cell, it samples three response modes: skeptic, balanced, and enthusiast. Each mode is prompted with the matched persona, the stratum context, the target concept, and the respondent's own open-ended answers on the other two concepts. The target concept's human rationale and outcome are never shown.

\paragraph{\textbf{Aggregation.}}
Each voice produces two samples. All six samples are scored with SSR. The primary aggregate averages the six PMFs:
\[
    \hat{p}(\Y\mid \D,\K,\X)=
    \frac{1}{|\mathcal{V}|}\sum_{v\in\mathcal{V}}\hat{p}_v(\Y\mid \D,\K,\X),
    \quad
    \mathcal{V}=\{\mathrm{skeptic},\mathrm{balanced},\mathrm{enthusiast}\}.
\]
We also evaluate selector checks. Oracle selection estimates how much useful signal exists in the voice pool. Learned selectors ask whether that signal can be recovered from deployable features.

\subsection{Evaluation}

\paragraph{\textbf{Human reason model evaluation.}}
The first evaluation asks whether human rationale-derived reasons help predict human behavior. We compare $\D,\K,\X$ against $\D,\K,\X+\Zcore$, and also compare $\D,\K,\X+\M$ against $\D,\K,\X+\M+\Zcore$. This tests whether reason states add signal beyond observed covariates and measured intermediate ratings.

\paragraph{\textbf{Reason-path checks.}}
For each core reason node $j$, we set that node to $+1$ and to $-1$ inside the learned readout:
\begin{equation}
    \Delta_j =
    \mathbb{E}\left[f_\theta(\D,\K,\X,\Z_{-j},\Z_j=+1)\right]
    -
    \mathbb{E}\left[f_\theta(\D,\K,\X,\Z_{-j},\Z_j=-1)\right].
    \label{eq:doz}
\end{equation}
Because the codebook defines $+1$ as support and $-1$ as opposition, $\Delta_j$ should be positive. This is an internal consistency check for the emulator, not proof of a natural causal effect.

\paragraph{\textbf{LLM simulation evaluation.}}
The second evaluation asks whether LLM-simulated reason states can replace human rationale-derived reasons. We evaluate $\Zhat$ directly against human $\Z$ and indirectly by passing $\Zhat$ through the human reason model. The no-leak reason generator receives only $\D,\K,\X$ and the codebook. It does not see human open text, purchase intent, believability, differentiation, rank, final choice, or gold reason labels.

\paragraph{\textbf{Controls for $\Zhat$.}}
We use two negative controls. In the no-reason control, all reason states are set to zero. In the mismatched-reason control, human $\Z$ vectors are shuffled across rows; this preserves the overall frequency of reasons but breaks the respondent-concept match. A useful LLM-simulated $\Zhat$ should improve over both controls.

\paragraph{\textbf{Native PMF versus $\Zhat$ readout.}}
For survey simulations there are two different predictions. The \emph{native PMF} is the simulator's own purchase-intent distribution, obtained from its generated text through SSR. The \emph{$\Zhat$ readout} first extracts signed reason states from the generated text and then predicts purchase intent with the human reason model $f_\theta(\D,\K,\X,\Zhat)$. The native PMF tests outcome fit; the $\Zhat$ readout tests whether the LLM-simulated reasons work in the human reason model.

\paragraph{\textbf{Metrics.}}
For purchase intent, lower MAE, $\frac{1}{n}\sum_r|\Y_r-\Yhat_r|$, and lower RMSE are better. Top-two-box intent is $T_r=\mathbb{1}[\Y_r\ge4]$; Top2 AUC is the ROC-AUC of $\Yhat$ for predicting $T$. Final-pick AUC is the ROC-AUC for whether a concept was the respondent's final choice. For reasons, we use active-set overlap, $\mathrm{IoU}=|A_{\mathrm{human}}\cap A_{\mathrm{model}}|/|A_{\mathrm{human}}\cup A_{\mathrm{model}}|$.

\section{Results}

\subsection{Human Reason Model Evaluation}

Human rationale-derived $\Zcore$ substantially improves held-out prediction (Table~\ref{tab:main}). The baseline model using only $\D,\K,\X$ has MAE 0.863. Adding $\Zcore$ lowers MAE to 0.625. Adding $\Zcore$ on top of believability and differentiation also helps, lowering MAE from 0.717 to 0.564. Respondent-level bootstrap intervals support the same lift; Appendix~\ref{app:experiments} reports the intervals.

\begin{table}[t]
    \centering
    \small
    \begin{tabular}{lrrrr}
        \toprule
        Model & MAE & Spearman & Top2 AUC & Final-pick AUC \\
        \midrule
        $\D,\K,\X$ & 0.863 & 0.245 & 0.628 & 0.602 \\
        $\D,\K,\X+\Zcore$ & 0.625 & 0.678 & 0.882 & 0.726 \\
        $\D,\K,\X+\M$ & 0.717 & 0.538 & 0.762 & 0.703 \\
        $\D,\K,\X+\M+\Zcore$ & \second{0.564} & \second{0.749} & \second{0.902} & \second{0.762} \\
        $\D,\K,\X+\M+\Z_{\mathrm{all}}$ & \best{0.535} & \best{0.772} & \best{0.917} & \best{0.779} \\
        \bottomrule
    \end{tabular}
    \caption{Main readout results. Best is bold; second-best is underlined. $\Z_{\mathrm{all}}$ includes outcome-near labels and is only a sensitivity check. The main claim uses $\Zcore$.}
    \label{tab:main}
\end{table}

\begin{figure}[t]
    \centering
    \includegraphics[width=0.98\linewidth]{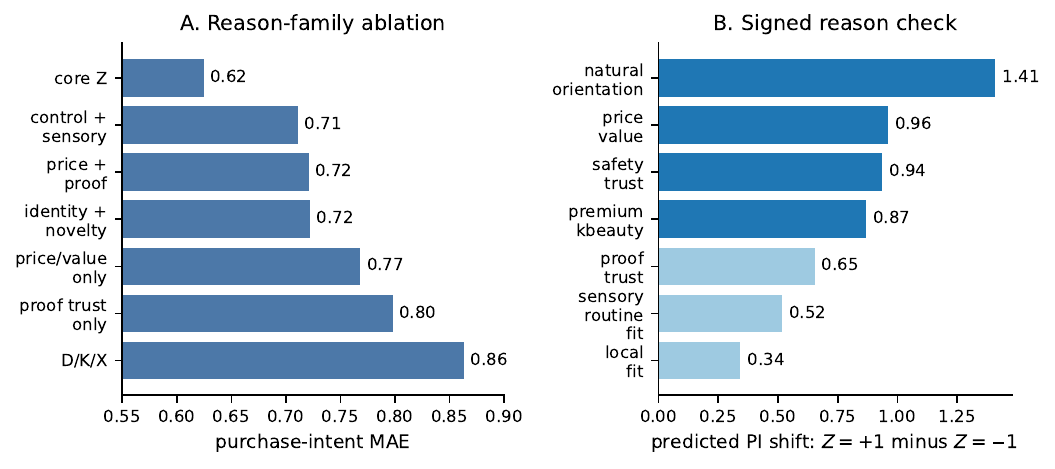}
    \caption{Reason-state checks for the human reason model. Left: ablation compares readouts that use different subsets of reason families. Right: each core reason node is set from blocker ($Z=-1$) to supporter ($Z=+1$); positive values mean the predicted purchase intent moves in the codebook-consistent direction.}
    \label{fig:mediators}
\end{figure}

The signed reason states also behave coherently inside the human reason model (Figure~\ref{fig:mediators}). When a core reason is set from blocker ($Z=-1$) to supporter ($Z=+1$), the predicted purchase intent increases. This does not identify a natural mediator effect, but it checks that the signed codebook and learned readout are aligned.

Reason states do not determine behavior by themselves. Two rows can share the same active reasons and still differ because the concept and respondent context differ. Appendix~\ref{app:surface} shows that the same reason family can have different strength across concepts. This is important: $\Z$ is not merely a disguised label for $\Y$.

\subsection{LLM Simulation Evaluation}

\begin{figure}[t]
    \centering
    \includegraphics[width=0.98\linewidth]{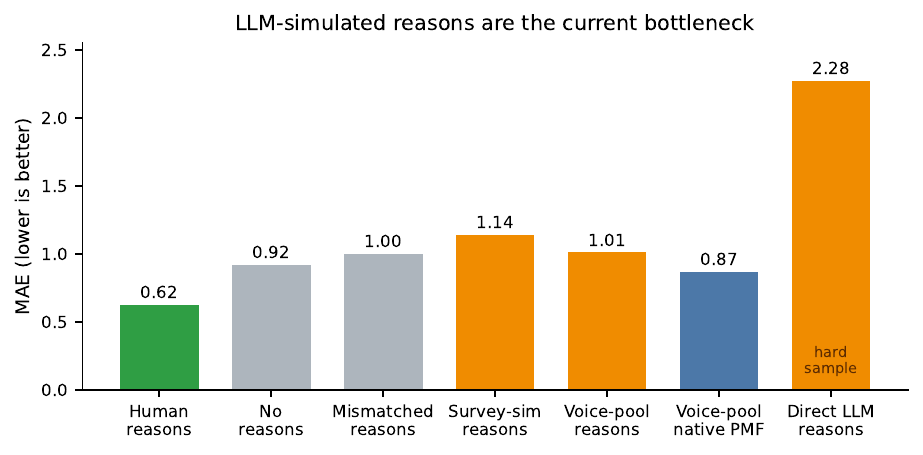}
    \caption{LLM reason simulation bottleneck. Green: human reason states passed through the human reason model. Gray: no-reason and mismatched-reason controls. Orange: LLM-simulated $\Zhat$ decoded through the same human reason model. Blue: the voice-pool simulator's native PMF, which predicts $\Y$ directly and does not test the $\Zhat\rightarrow\Y$ path. The direct LLM reason bar is evaluated on 30 hard rows.}
    \label{fig:bottleneck}
\end{figure}

The main negative result is that LLM-simulated reason states do not yet substitute for human rationale-derived reasons. Figure~\ref{fig:bottleneck} separates outcome simulation from reason-state simulation. The human-reasons bar uses the observed rationale-derived $\Z$. The no-reason control sets all $\Z$ values to zero. The mismatched-reasons control shuffles human $\Z$ across respondent-concept rows. Survey-simulated and voice-pool reasons are $\Zhat$ states extracted from LLM survey text and decoded through the human reason model. The voice-pool native PMF is different: it is the voice-pool simulator's direct purchase-intent prediction through SSR, without passing through $\Zhat$.

Human reasons perform best. LLM-simulated reasons often hurt when decoded through the same human reason model, and the direct no-leak reason generator is worse than both no-reason and mismatched-reason controls on the hard sample. The common failure mode is concept echoing. The LLM reads the concept board and reproduces the intended selling points, even when a respondent rejects those same points. For example, one human rationale expressed skepticism that plant-based ingredients could provide adequate protection. The human code marks proof trust and natural orientation as blockers. An LLM-simulated reason may instead mark natural orientation and local fit as supports because the concept presents those cues positively.

The voice-pool results show a capacity gap. Oracle voice selection is much better than uniform aggregation, suggesting that useful modes exist in the pool. However, learned selectors using only deployable features remain close to uniform aggregation. The bottleneck is therefore not only generation; it is also selecting the appropriate response mode for a given respondent-concept cell.

\section{Discussion and Limitations}

Outcome matching is an incomplete validation target for social simulation. A simulator can match a winner or an aggregate distribution while missing the reason path that would make the result actionable. The sunscreen concept test shows that human rationale-derived reasons improve held-out prediction, move predictions in the expected direction under signed perturbations, and expose concept-level heterogeneity. LLM-simulated reasons do not yet pass the same test: they often echo the concept board rather than recover the respondent's acceptance or rejection path.

The causal claim is intentionally limited. We do not estimate natural direct or indirect effects, and the study is observational with respect to stated reasons. The contribution is an evaluation object: a fixed reason codebook and a fixed human reason model create a manipulable interface for prediction, perturbation, and simulator substitution. The study is small and covers one category. Future work should repeat the protocol in other categories, pre-register the codebook, and compare stronger LLM reason generators, retrieval, and learned voice selectors.

\section{Conclusion}

Matching human answer distributions is not enough for social simulation. A simulator can predict the winner and still miss the mechanism. We propose a reason-mediated behavioral model: $\D,\K,\X$ shape $\Z$, and $\Z$ helps predict $\Y$. In one sunscreen concept test, human rationale-derived reasons improve prediction. Current LLM-simulated reasons do not recover the same reason path reliably. Reasons are therefore not merely explanatory text; they are part of what a simulator should be evaluated against.

\section*{Acknowledgments}

This work was supported in part by a grant from Exception Raised.

\bibliography{causal_emulator_colm2026}
\bibliographystyle{colm2026_conference}

\clearpage
\appendix

\section{Reproducibility and Ethics}

\paragraph{Reproducibility.}
All results were computed from the local sunscreen concept-test workspace: human survey rows, fixed reason-code files, processed feature matrices, LLM-simulated reason checks, and figure scripts. This draft is a new LaTeX/PDF copy, not an overwrite of the previous paper.

\paragraph{Ethics.}
Synthetic respondents can mislead users if they are treated as replacements for human evidence. This paper argues for a stricter test. The study is small, and the results should not be used as population claims without more validation.

\section{Concept-test Questionnaire and Stimuli}
\label{app:study-details}

\paragraph{Target and sample.}
The concept test targeted Indian women aged 22--40 who had purchased sunscreen in the past six months. The fielding brief specified English-speaking, urban or semi-urban respondents with an NCCS A/B skew. After exclusions, the analytic file contains 94 usable respondents. Six responses were excluded for attention, rurality, or straight-line quality rules.

\begin{table}[h]
    \centering
    \small
    \begin{tabular}{p{0.34\linewidth}p{0.56\linewidth}}
        \toprule
        Quantity & Observed distribution \\
        \midrule
        Respondents & 94 usable; 6 excluded before analysis \\
        Age & range 22--40; mean 28.7; median 28 \\
        City tier & Tier 1: 51; Tier 2: 19; Tier 3: 24 \\
        Concept order & A,B,C: 23; A,C,B: 11; B,A,C: 20; B,C,A: 12; C,A,B: 19; C,B,A: 9 \\
        Sunscreen-use frequency code & 1: 58; 2: 21; 3: 5; 4: 6; 5: 4 \\
        White-cast concern & ratings 4--5: 80 / 94 \\
        SPF-claim trust & ratings 1--5: 2, 14, 46, 22, 10 \\
        Final pick code & A: 32; B: 39; C: 19; none: 4 \\
        \bottomrule
    \end{tabular}
    \caption{Sample and design distribution. Code meanings are preserved from the survey export when the raw label map is not present in the analysis file.}
    \label{tab:study-distribution}
\end{table}

\paragraph{Questionnaire fields.}
The file contains age, city tier, education, chief-earner education, household durables, income, languages, BFI-10 personality items, sunscreen-use frequency, sunscreen brands owned, white-cast concern, SPF-claim trust, purchase intent, believability, differentiation, open-ended rationales, rankings, and final pick.

\begin{table}[h]
    \centering
    \small
    \begin{tabular}{p{0.16\linewidth}p{0.36\linewidth}p{0.38\linewidth}}
        \toprule
        Symbol & Survey source & Examples \\
        \midrule
        $\D_i$ & Respondent descriptors & age, city tier, education, income proxy, languages, BFI-10 personality items \\
        $\K_i$ & Sunscreen-category context & use frequency, brands owned, white-cast concern, SPF-claim trust \\
        $\X_c$ & Concept board & price, origin, authority cue, claims, proof points, sensory promise \\
        $\M_{ic}$ & Measured intermediate ratings & believability and differentiation for concept $c$ \\
        $\Z_{ic}$ & Open-ended rationale after concept $c$ & signed reason states extracted from the respondent's text \\
        $\Y_{ic}$ & Behavioral response & purchase intent; ranking and final pick as secondary outcomes \\
        \bottomrule
    \end{tabular}
    \caption{How the questionnaire maps to the variables in the reason-mediated behavioral model.}
    \label{tab:survey-mapping}
\end{table}

\begin{figure}[h]
    \centering
    \includegraphics[width=0.46\linewidth]{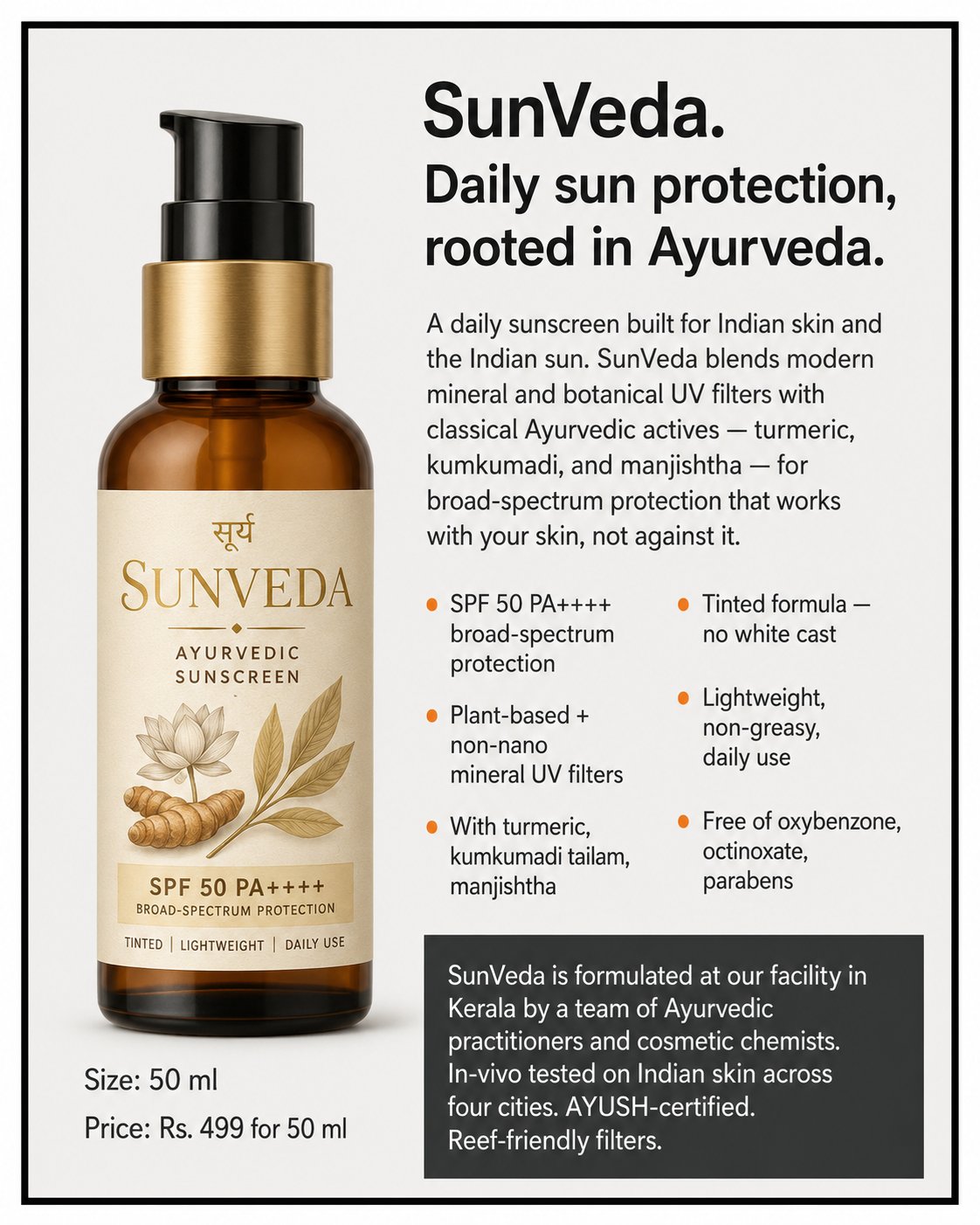}
    \caption{Example concept board shown to respondents. The full study used three boards: SunVeda, Dr. Anita, and Hae.}
    \label{fig:concept-example}
\end{figure}

\paragraph{Scale anchors.}
Purchase intent used a 1--5 scale: 1 definitely would not buy; 2 probably would not buy; 3 might or might not buy; 4 probably would buy; 5 definitely would buy. Believability used 1 not at all believable to 5 extremely believable. Differentiation used 1 not at all different to 5 extremely different.

\begin{table}[h]
    \centering
    \scriptsize
    \setlength{\tabcolsep}{3pt}
    \begin{tabular}{>{\raggedright\arraybackslash}p{0.14\linewidth}>{\raggedright\arraybackslash}p{0.18\linewidth}>{\raggedright\arraybackslash}p{0.60\linewidth}}
        \toprule
        Concept & Price / positioning & Stimulus summary \\
        \midrule
        A: SunVeda & Rs. 499 / 50 ml; Ayurvedic-botanical & ``Daily sun protection, rooted in Ayurveda.'' Daily sunscreen for Indian skin and Indian sun; mineral and botanical UV filters with turmeric, kumkumadi, and manjishtha. Benefits included SPF 50 PA++++, tinted no-white-cast formula, lightweight non-greasy daily use, and avoidance of oxybenzone, octinoxate, and parabens. Proof included Kerala formulation, Indian-skin testing, AYUSH certification, and reef-friendly filters. \\
        B: Dr. Anita & Rs. 699 / 50 ml; dermatologist-clinical Indian & ``Dr. Anita Invisible SPF 50. Formulated for Indian skin. Zero white cast.'' Dermatologist-formulated sunscreen for Indian skin tones, heat, humidity, and pollution. Benefits included SPF 50 PA++++, invisible finish, sweat/humidity resistance, matte finish, non-comedogenic use, and face/neck daily use. Proof included Indian dermatologist formulation, a 12-week test on 142 Indian women, 91\% no visible white cast, and BIS-tested SPF/PA ratings. \\
        C: Hae & Rs. 1,299 / 50 ml; Korean prestige hybrid & ``Hae. K-beauty sun fluid for the Indian sun.'' Korean-formulated, India-available lightweight sun fluid with next-generation Korean UV filters, barely-there feel, and subtle tone-up tint. Benefits included SPF 50+ PA++++, watery texture, primer use, vegan, fragrance-free, and alcohol-free claims. Proof included Seoul formulation/manufacture, Indian distribution, non-US-approved filter caveat, Seoul clinical testing, and Mumbai compatibility check. \\
        \bottomrule
    \end{tabular}
    \caption{Concept stimuli used in the three-arm sunscreen concept test.}
    \label{tab:concept-stimuli}
\end{table}

\clearpage

\section{Experiment Details}
\label{app:experiments}

\paragraph{Rows and artifacts.}
The unit is a respondent-concept row. The concept test contains 94 respondents and three concepts, giving 282 rows. The artifact set includes the human survey matrix, fixed reason-code files, processed feature matrices, synthetic response outputs, voice-pool outputs, a hard no-leak staged-generator sample, and LLM-readout checks.

\paragraph{Human-reason-model experiments.}
We train purchase-intent readouts using five-fold GroupKFold by respondent. The headline model is ridge regression; final-pick checks use balanced logistic regression. Robustness checks include respondent bootstrap intervals, concept-held-out splits, zero-$\Z$ controls, shuffled-$\Z$ controls, and reason-family ablations. We also force each core reason node to $-1,0,+1$ and measure readout movement.

\paragraph{Reason-surface experiments.}
We group rows by signed reason signatures and inspect the mapping from reason states to purchase intent across concepts. This asks whether the same reason family has stable direction while still allowing concept-level heterogeneity. It also checks that $\Z$ is not merely a deterministic rewrite of $\Y$.

\begin{figure}[h]
    \centering
    \includegraphics[width=0.70\linewidth]{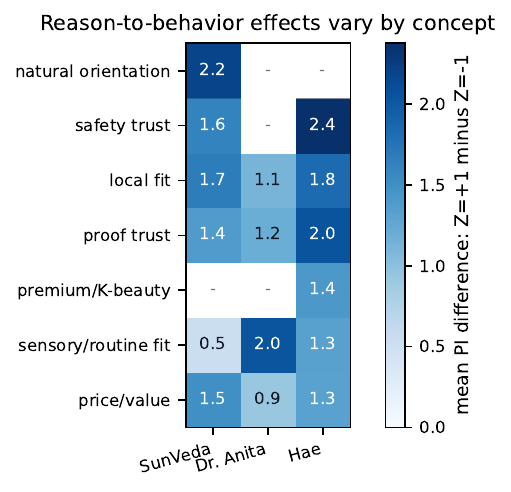}
    \caption{Observed reason-to-behavior associations by concept. Each cell is the mean purchase-intent difference between rows where a reason supports purchase ($Z=+1$) and rows where it blocks purchase ($Z=-1$). Blank cells mean that this small study did not contain both sides of the contrast.}
    \label{fig:surface}
\end{figure}

\paragraph{Reason-surface plot.}
\label{app:surface}
Figure~\ref{fig:surface} reports the observed reason-to-behavior contrasts by concept.

\paragraph{LLM-simulated reason experiments.}
We replace human $\Z$ with $\Zhat$ from the context-conditioned response simulation, the voice-pool simulation, and a staged no-leak LLM generator. Each $\Zhat$ is decoded through the same human reason model. We report reason IoU, signed-node accuracy, active precision/recall, and human-reason-model MAE.

\paragraph{Voice-selection experiments.}
For the voice-pool simulation, we compare the current average over skeptic/balanced/enthusiast samples with row-level oracle selection, sample-level oracle selection, and learned hard/soft selectors. The oracle results measure available capacity; learned selectors measure deployability.

\paragraph{Transition and warm-start experiments.}
We also train exploratory transition models from observed respondent/category state to reason state, $\D,\K,\X\rightarrow\Z$, and warm-start models that use the same respondent's other-concept reasons to predict the held-out concept. These are exploratory because rare reason nodes are hard to recover in a 94-person study.

\paragraph{Bootstrap intervals.}
For the clean reason readout, $\D,\K,\X+\Zcore$ improves over $\D,\K,\X$ by 0.239 MAE with 95\% CI $[0.172,0.299]$. Adding $\Zcore$ on top of $\M$ improves over $\D,\K,\X+\M$ by 0.154 MAE with 95\% CI $[0.099,0.209]$.

\paragraph{Hard-sample LLM reason metrics.}
The no-leak staged generator was evaluated on 30 difficult rows. Mean reason IoU was 0.203; exact active-set match was 0.000; signed-node accuracy was 0.448; active recall was 0.625; active precision was 0.210. Human-reason-model MAE was 1.278 for human $\Z$, 2.276 for staged LLM $\Zhat$, 1.760 for zero-$\Z$, and 1.853 for shuffled-$\Z$.

\section{Models and Readouts}
\label{app:models}

\paragraph{Human reason model.}
The human reason model is the readout $f_\theta(\D,\K,\X,\Z)\rightarrow\Y$. The primary model is ridge regression with $\alpha=1.0$, median imputation for numeric variables, most-frequent imputation for categoricals, standard scaling for numeric variables, and one-hot encoding for categorical concept features. Predictions are clipped to 1--5.

\paragraph{Other readouts.}
Gradient boosting and random forest readouts were used as nonlinear robustness checks. They test whether the reason-state lift depends on using a linear readout.

\paragraph{Prompted LLM readout.}
The prompt-only LLM readout used \texttt{gpt-5.4-mini} on a stratified sample. One arm used reason states only; another used $\D,\K,\X,\M,\Z$ plus a verbalized structured-model prior. These readouts did not beat the structured readout.

\paragraph{LLM reason generator.}
The staged no-leak $\Zhat$ generator used the OpenAI verifier/generator pipeline. The reported hard-sample run uses a GPT-5-family structured generator; the same family is used for larger-model verification in the human $\Z$ extraction pipeline. This choice is deliberate: $\D,\K,\X\rightarrow\Zhat$ generation is a smaller, structured JSON task, while the Gemma runs are high-volume respondent-style survey simulations. The generator saw only $\D,\K,\X$ and the reason codebook. It did not see open text, outcomes, or gold reason labels.

\paragraph{LLM survey-simulation models.}
Synthetic responses use Gemma 4 31B IT for response generation and Qwen3-Embedding-0.6B for semantic-similarity rating embeddings. These Gemma outputs are used in two ways: as native outcome predictions through SSR, and as generated survey text from which $\Zhat$ can be extracted for the $\Zhat\rightarrow\Y$ test. Direct-Likert arms produce a single integer. SSR arms produce a short free-text answer, which is converted into a 1--5 purchase-intent probability mass function using six locked reference-statement sets. Appendix~\ref{app:sim-arms} gives the full arm table.

\paragraph{Persona matching.}
Nemotron-Personas-India rows are filtered to female, urban personas aged 22--40. Each human respondent is matched to one persona by seeded stratified matching on city tier and age band. The primary match bucket is exact city tier and exact age band; fallback buckets use same city tier or same age band. Assigned personas are removed from the pool, so the matching is without replacement.

\paragraph{Context-SSR details.}
The Context-SSR arm combines the matched Nemotron persona with a stratum-level sunscreen context block. The context block is generated from non-outcome survey fields: sunscreen-use frequency, brands owned, white-cast concern, SPF-claim trust, languages, and personality summaries. It excludes purchase intent, believability, differentiation, rankings, final choice, and open-ended product rationales.

\section{Simulation Arm Taxonomy}
\label{app:sim-arms}

\paragraph{Arm taxonomy.}
The simulation run contains eight arms. They form a controlled ladder from demographic-only prompting to persona, psychographic, category-context, and own-answer conditioning. The reason-path claims use the LLM reason simulation protocols described in Section~4; the full arm list is included for reproducibility.

\begin{table}[h]
    \centering
    \small
    \setlength{\tabcolsep}{4pt}
    \begin{tabular}{lll}
        \toprule
        Arm & Name used here & Conditioning \\
        \midrule
        A1 & Demo-DLR & demographics only; direct 1--5 rating \\
        A2 & Demo-SSR & demographics only; free text scored by SSR \\
        N0 & Persona-DLR & Nemotron persona; direct 1--5 rating \\
        N1 & Persona-SSR & Nemotron persona; free text scored by SSR \\
        N2a & OCEAN-Numerical & persona plus numeric Big Five scores \\
        N2b & OCEAN-Narrative & persona plus prose psychographic narrative \\
        N3 & Context-SSR & persona plus stratum-level sunscreen context \\
        N4 & Half-Q & persona plus own non-target survey answers \\
        \bottomrule
    \end{tabular}
    \caption{Simulation arms. DLR is direct Likert rating. SSR is semantic similarity rating: the model writes text, and an embedding-based rater maps the text to a 1--5 probability mass function.}
    \label{tab:sim-arms}
\end{table}

\paragraph{Generation scale.}
The bulk run covers 94 respondents, three concepts, eight arms, and three samples per arm-cell, giving $94\times3\times8\times3=6{,}768$ generated responses. Response generation uses Gemma 4 31B IT through a local OpenAI-compatible endpoint. SSR embeddings use Qwen3-Embedding-0.6B.

\section{Semantic Similarity Rating}
\label{app:ssr}

\paragraph{Text-to-PMF scoring.}
For SSR arms, the model output is not parsed as a number. It is embedded and compared with six locked sets of reference statements covering the 1--5 purchase-intent scale. The final probability mass function is the average across reference sets:
\[
    \hat{p}(\Y=k \mid r)=
    \frac{1}{6}\sum_{s=1}^{6}
    \hat{p}_s(\Y=k \mid \mathrm{embed}(r)),
    \quad k\in\{1,\ldots,5\}.
\]
Mean purchase intent is then $\sum_{k=1}^{5} k\hat{p}(\Y=k)$, and top-two-box mass is $\hat{p}(\Y=4)+\hat{p}(\Y=5)$. In the reason-path experiments, the native PMF is used as an outcome prediction; the generated text is separately mapped into $\Zhat$ for the human-reason-model test.

\section{Reason Codebook}
\label{app:codebook}

\begin{table}[h]
    \centering
    \small
    \begin{tabular}{p{0.30\linewidth}p{0.27\linewidth}p{0.34\linewidth}}
        \toprule
        Core node & General family & Positive / negative interpretation \\
        \midrule
        \texttt{cz\_price\_value} & Value/control & worth the price / too expensive or poor value \\
        \texttt{cz\_proof\_trust} & Proof trust & evidence is believed / claims are doubted \\
        \texttt{cz\_natural\_orientation} & Natural/cultural fit & natural or traditional cue helps / cue reduces trust \\
        \texttt{cz\_premium\_kbeauty} & Novelty/status & premium or K-beauty appeal / feels excessive or irrelevant \\
        \texttt{cz\_local\_fit} & Local fit & Indian skin/climate fit helps / local fit is missing \\
        \texttt{cz\_sensory\_routine\_fit} & Routine fit & texture/no-white-cast helps daily use / routine fit blocks adoption \\
        \texttt{cz\_safety\_trust} & Safety trust & safety is reassured / safety or regulatory anxiety blocks adoption \\
        \bottomrule
    \end{tabular}
    \caption{Core reason codebook used for $\Zcore$.}
    \label{tab:codebook}
\end{table}

\paragraph{Human reason extraction prompt excerpt.}
\begin{quote}\small
\texttt{You extract respondent-stated reason states for a sunscreen concept test. Return only JSON. Do not infer hidden psychology from the rating, rank, or final pick; those are not provided. Use the concept context only to disambiguate words like price, Korean, dermatologist, SPF, or Indian skin. Classify only reasons that are supported by the respondent's own open-ended text. Do not mark a node just because the concept has that attribute.}
\end{quote}

\paragraph{No-leak reason generation prompt excerpt.}
\begin{quote}\small
\texttt{Task: predict reason states Z\_hat for one respondent-concept cell using only respondent state D, category prior K, and concept treatment X. You do not see human open text, gold reason labels, ratings, ranking, or final choice. Use a two-stage reasoning process internally: candidate reason discovery, then calibrated compression into a sparse signed Z\_hat vector. Most rows should have 0--2 active reason families; use 3 only for a clear mixed tradeoff. Do not activate a generic concept attribute unless D/K suggests the respondent will use it as a decision reason.}
\end{quote}

\section{Voice-pool Aggregation}
\label{app:polyvoice}

\paragraph{Algorithm.}
The voice-pool estimator keeps several possible response modes alive before averaging. In this version the three modes are skeptic, balanced, and enthusiast. Each voice produces a candidate reason state and a purchase-intent distribution. The selector weights them:
\[
    \hat{p}(\Y\mid \D,\K,\X)=\sum_{v\in\mathcal{V}}\pi_v(\D,\K,\X)\,\hat{p}_v(\Y\mid \D,\K,\X).
\]

\begin{figure}[h]
    \centering
    \includegraphics[width=0.96\linewidth]{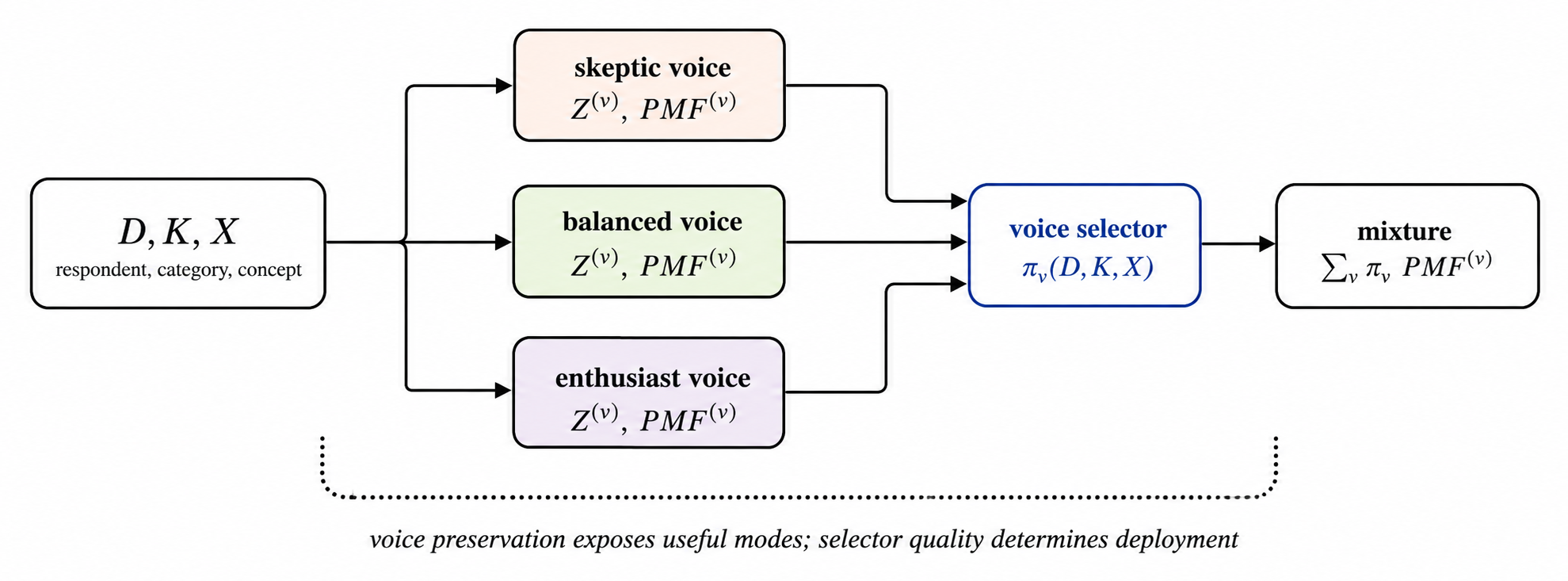}
    \caption{Voice-pool aggregation keeps skeptic, balanced, and enthusiast modes separate before mixing their predictions.}
    \label{fig:polyvoice}
\end{figure}

\paragraph{Selector tests.}
For each respondent-concept cell, uniform aggregation was compared with a row-level oracle voice, a sample-level oracle, and learned hard/soft selectors. The large oracle gap means the pool contains useful modes. The weak learned selector means selecting the right mode is still unsolved.

\clearpage
\section{Additional Result Tables}
\label{app:additional-tables}

\begin{table}[H]
    \centering
    \small
    \begin{tabular}{lrrrr}
        \toprule
        Readout & N & MAE & RMSE & Top2 AUC \\
        \midrule
        human $\Z$, human reason model & 282 & \best{0.625} & \best{0.837} & \best{0.882} \\
        zero-$\Z$ control & 282 & 0.918 & 1.103 & 0.566 \\
        shuffled-$\Z$ control & 282 & 1.001 & 1.232 & 0.518 \\
        survey-simulated $\Zhat$, human reason model & 282 & 1.140 & 1.437 & 0.511 \\
        voice-pool $\Zhat$, human reason model & 282 & 1.013 & 1.285 & 0.618 \\
        voice-pool native PMF & 282 & \second{0.867} & \second{1.035} & \second{0.652} \\
        \bottomrule
    \end{tabular}
    \caption{LLM-simulated reason and native-PMF comparisons. Best is bold; second-best is underlined.}
    \label{tab:synthetic-readouts}
\end{table}

\begin{table}[h]
    \centering
    \small
    \begin{tabular}{lrr}
        \toprule
        Source & Current MAE & Oracle voice MAE \\
        \midrule
        generic & \best{0.867} & \second{0.485} \\
        context & \second{0.930} & 0.517 \\
        demographic-context & 0.935 & \best{0.476} \\
        persona & 0.932 & 0.495 \\
        \bottomrule
    \end{tabular}
    \caption{Voice-pool capacity check. The oracle column is not deployable; it estimates possible gain if voice selection were solved.}
    \label{tab:voice-capacity}
\end{table}

\section{Additional Robustness Checks}
\label{app:robustness-checks}

This section reports additional checks computed after the main submission. They are included for camera-ready transparency and do not change the main-paper tables or figures.

\paragraph{Text and valence baselines.}
Table~\ref{tab:text-valence-baselines} adds two simpler rationale-derived baselines. The raw-text baseline uses TF-IDF features from the open-ended rationale. The valence baseline compresses the rationale into signed positive/negative valence counts. These baselines are not deployable as no-leak simulators because they use the human open text, but they help interpret the main result: part of the predictive value of $\Z$ comes from signed valence, and part comes from the structured reason codebook.

\begin{table}[h]
    \centering
    \small
    \begin{tabular}{lrrrr}
        \toprule
        Model & MAE & Spearman & Top2 AUC & Final-pick AUC \\
        \midrule
        $\D,\K,\X$ & 0.863 & 0.245 & 0.628 & 0.602 \\
        $\D,\K,\X+\Zcore$ & 0.625 & 0.678 & 0.882 & 0.726 \\
        $\D,\K,\X+\M+\Zcore$ & 0.564 & \best{0.749} & \best{0.902} & 0.762 \\
        $\D,\K,\X+$ valence & 0.593 & 0.674 & 0.873 & 0.746 \\
        $\D,\K,\X+$ raw-text TF-IDF & 0.765 & 0.466 & 0.774 & 0.711 \\
        $\D,\K,\X+\M+$ raw-text TF-IDF & 0.663 & 0.655 & 0.847 & 0.730 \\
        $\D,\K,\X+\M+$ valence & \best{0.563} & 0.735 & 0.890 & \best{0.777} \\
        \bottomrule
    \end{tabular}
    \caption{Additional human-open-text baselines. Lower MAE is better; higher Spearman and AUC are better. These baselines use human rationale text and are therefore diagnostic baselines, not no-leak simulators.}
    \label{tab:text-valence-baselines}
\end{table}

\paragraph{Full-sample reason agreement.}
Table~\ref{tab:reason-agreement-extra} reports reason-state agreement between human rationale-derived $\Z$ and generated $\Zhat$. IoU measures overlap between active reason sets. Signed accuracy measures whether each signed reason node is matched exactly. Active precision asks whether generated active reasons were also active in the human rationale; active recall asks whether human active reasons were recovered. The oracle rows are capacity checks: they select the best available generated sample with respect to the human reason state and are not deployable.

\begin{table}[h]
    \centering
    \small
    \begin{tabular}{lrrrrr}
        \toprule
        Method & N & IoU & Signed acc. & Active prec. & Active rec. \\
        \midrule
        Gemma Context-SSR majority & 282 & 0.201 & 0.569 & 0.236 & 0.442 \\
        Voice-pool majority & 282 & 0.219 & 0.458 & 0.231 & 0.627 \\
        Gemma Context-SSR oracle & 282 & 0.281 & 0.681 & 0.336 & 0.463 \\
        Voice-pool oracle & 282 & \best{0.471} & \best{0.767} & \best{0.496} & \best{0.727} \\
        OpenAI no-leak hard sample & 30 & 0.203 & 0.448 & 0.210 & 0.625 \\
        \bottomrule
    \end{tabular}
    \caption{Additional reason-state agreement checks. The oracle rows estimate capacity in the generated sample pool; they are not deployable model results.}
    \label{tab:reason-agreement-extra}
\end{table}

\paragraph{Bootstrap intervals for reason agreement.}
Table~\ref{tab:reason-ci-extra} gives respondent-bootstrap 95\% intervals for the same reason agreement metrics.

\begin{table}[h]
    \centering
    \scriptsize
    \begin{tabular}{lrrrr}
        \toprule
        Method & IoU & Signed acc. & Active prec. & Active rec. \\
        \midrule
        Gemma Context-SSR majority & .201 [.175,.229] & .569 [.549,.588] & .236 [.202,.270] & .442 [.396,.492] \\
        Voice-pool majority & .219 [.192,.248] & .458 [.436,.480] & .231 [.204,.261] & .627 [.586,.666] \\
        Gemma Context-SSR oracle & .281 [.249,.316] & .681 [.661,.702] & .336 [.295,.383] & .463 [.422,.510] \\
        Voice-pool oracle & .471 [.435,.510] & .767 [.748,.785] & .496 [.453,.537] & .727 [.689,.766] \\
        OpenAI no-leak hard sample & .203 [.125,.291] & .448 [.378,.520] & .210 [.132,.299] & .625 [.452,.784] \\
        \bottomrule
    \end{tabular}
    \caption{Bootstrap intervals for generated reason agreement.}
    \label{tab:reason-ci-extra}
\end{table}

\paragraph{Sign confusion and prevalence calibration.}
Table~\ref{tab:sign-prevalence-extra} separates two failure modes. A low flip rate means that, when both human and generated reasons are active, the sign is usually aligned. A high prevalence error means that the generator activates too many or too few reason families overall. The generated methods still have substantial prevalence error, especially on sensory/routine fit and local fit.

\begin{table}[h]
    \centering
    \small
    \begin{tabular}{lrrrr}
        \toprule
        Method & Flip rate & Human + rec. & Human - rec. & Active-prev. error \\
        \midrule
        Gemma Context-SSR majority & 0.322 & 0.560 & 0.226 & 0.228 \\
        Voice-pool majority & 0.254 & 0.731 & 0.435 & 0.413 \\
        Gemma Context-SSR oracle & 0.225 & 0.566 & 0.274 & 0.160 \\
        Voice-pool oracle & \best{0.062} & \best{0.796} & \best{0.601} & \best{0.113} \\
        OpenAI no-leak hard sample & 0.333 & 0.909 & 0.385 & 0.452 \\
        \bottomrule
    \end{tabular}
    \caption{Sign confusion and prevalence calibration. Human + rec. is recall for positive human reasons; human - rec. is recall for negative human reasons.}
    \label{tab:sign-prevalence-extra}
\end{table}

\paragraph{Supervised ceiling for $\D,\K,\X\rightarrow\Z$.}
Table~\ref{tab:dkx-z-ceiling-extra} asks how well observed respondent/category state and concept features can predict reason states without human rationale text. The ceiling is modest for most reason families, which supports the underdetermination concern: exact individual reason states are not fully recoverable from $\D,\K,\X$ in this study. A stronger simulator should therefore represent uncertainty over $\Z$, not only a single deterministic reason vector.

\begin{table}[H]
    \centering
    \small
    \begin{tabular}{lrrrr}
        \toprule
        Reason node & Best model & Macro F1 & Active F1 & Entropy \\
        \midrule
        price/value & logistic & 0.491 & 0.579 & 0.583 \\
        proof trust & random forest & 0.480 & 0.548 & 0.866 \\
        natural orientation & logistic & \best{0.520} & \best{0.715} & 0.209 \\
        premium/K-beauty & random forest & 0.449 & 0.618 & 0.339 \\
        local fit & gradient boosting & 0.334 & 0.145 & 0.262 \\
        sensory/routine fit & gradient boosting & 0.443 & 0.261 & 0.293 \\
        safety trust & logistic & 0.359 & 0.135 & 0.382 \\
        \bottomrule
    \end{tabular}
    \caption{Supervised cold $\D,\K,\X\rightarrow\Z$ ceiling by reason node. Entropy is the mean predictive entropy of the best selected model.}
    \label{tab:dkx-z-ceiling-extra}
\end{table}

\end{document}